\pdfoutput=1

\documentclass[11pt]{article}

\usepackage[final]{acl}

\usepackage{times}
\usepackage{latexsym}

\usepackage[T1]{fontenc}

\usepackage[utf8]{inputenc}

\usepackage{microtype}
\usepackage{multirow}
\usepackage{inconsolata}
\usepackage{caption}
\usepackage{graphicx}
\usepackage{array}
\usepackage{subcaption}
\usepackage{booktabs}
\usepackage{dblfloatfix}

\usepackage{gb4e}
\noautomath
\usepackage{subcaption}

\title{Is linguistically-motivated data augmentation worth it?}

\author{Ray Groshan* \and Michael Ginn*  \and Alexis Palmer \\ University of Colorado \\ 
\textbf{Correspondence:} michael.ginn@colorado.edu  \\ 
\small $*$ Equal contribution
}

\begin{document}
\maketitle
\begin{abstract}
Data augmentation, a widely-employed technique for addressing data scarcity, involves generating synthetic data examples which are then used to augment available training data. Researchers have seen surprising success from simple methods, such as random perturbations from natural examples, where models seem to benefit even from data with nonsense words, or data that doesn't conform to the rules of the language. A second line of research produces synthetic data that does in fact follow all linguistic constraints; these methods require some linguistic expertise and are generally more challenging to implement. No previous work has done a systematic, empirical comparison of both linguistically-naive and linguistically-motivated data augmentation strategies, leaving uncertainty about whether the additional time and effort of linguistically-motivated data augmentation work in fact yields better downstream performance.

In this work, we conduct a careful and comprehensive comparison of augmentation strategies (both linguistically-naive and linguistically-motivated) for two low-resource languages with different morphological properties, Uspanteko and Arapaho. We evaluate the effectiveness of many different strategies and their combinations across two important sequence-to-sequence tasks for low-resource languages: machine translation and interlinear glossing. We find that linguistically-motivated strategies can have benefits over naive approaches, but only when the new examples they produce are not significantly unlike the training data distribution. 




\end{abstract}
\defcitealias{319093}{Cholaj Tzijb’al li Uspanteko}

\section{Introduction}
\begin{figure}
    \centering
    \includegraphics[width=\linewidth]{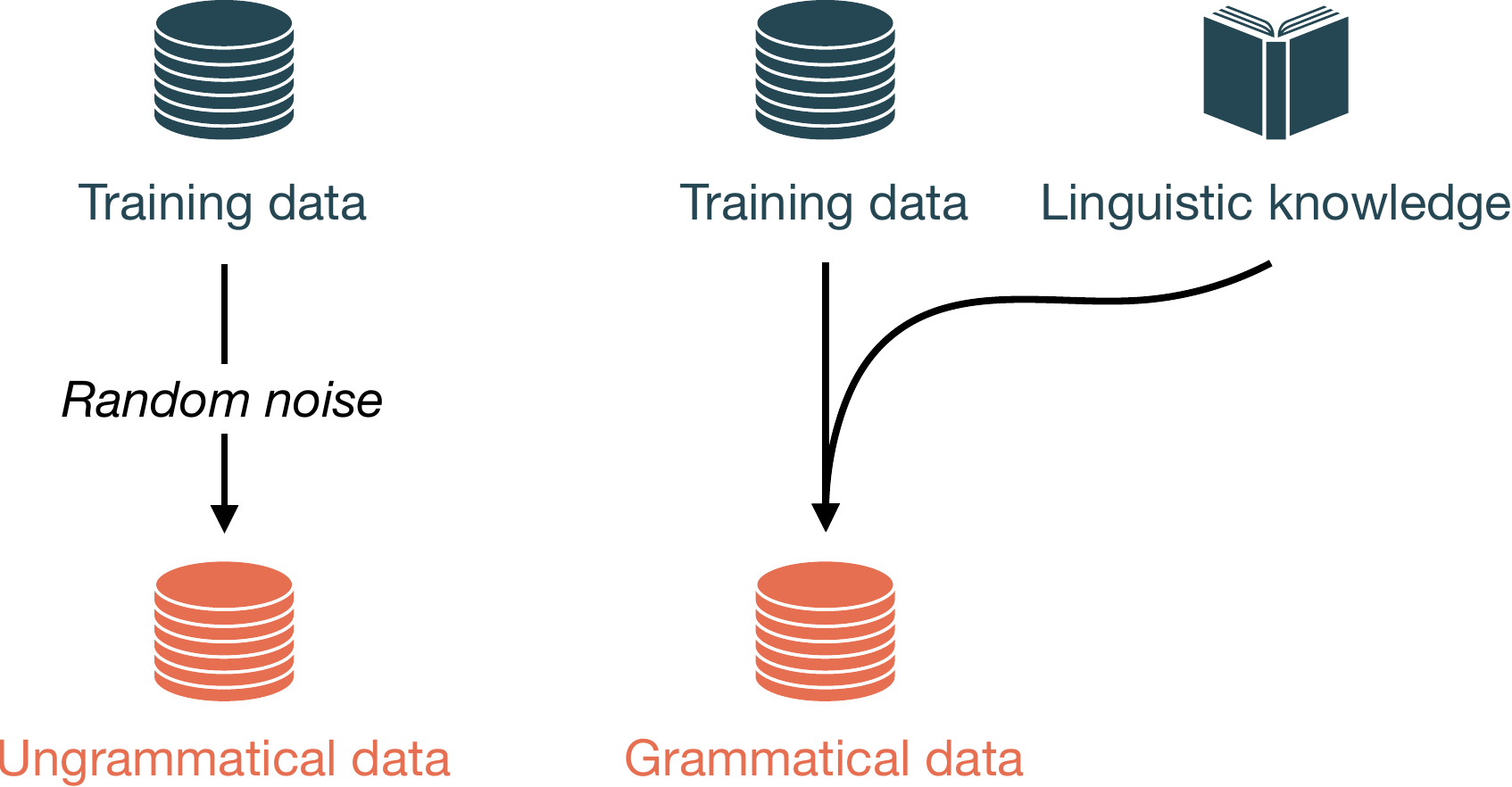}
    \caption{Two types of approach to data augmentation. 
    \textbf{Naive augmentation (left)} uses random perturbations to produce new examples which are not necessarily grammatically valid, while \textbf{linguistically-informed augmentation (right)} uses linguistic knowledge to constrain synthetic examples to be grammatically valid.}
    \label{fig:diagram}
\end{figure}
\textit{Data augmentation} refers to techniques that are used to create additional, artificial examples for training machine learning models in order to increase the total amount of training data. Data augmentation has been well-studied in computer vision, where simple perturbations  such as flipping, rotating, or recoloring images are applied on natural data \citep{gradient_lecun_1998}. Similar approaches that depend on random perturbation have been used in NLP for tasks such as morphological inflection \citep{silfverberg-etal-2017-data, bergmanis-etal-2017-training, anastasopoulos-neubig-2019-pushing, yang-etal-2022-generalizing}, classification \citep{Wei2019, karimi-etal-2021-aeda-easier}, and machine translation \citep{wang-etal-2018-switchout, guo-etal-2020-sequence}.

One limitation of such approaches is that they often create new examples which are not linguistically valid. For example, a strategy which randomly inserts words might produce an ungrammatical sentence such as "The dog chases \textit{bird} the cat," where \textit{bird} is the inserted word. To address this issue, some researchers leverage linguistic resources to produce examples that are both novel \emph{and} grammatical \citep{NIPS2015_250cf8b5, Wei2019, pratapa-etal-2018-language, seo-etal-2023-chef}.

Designing linguistically-motivated strategies (\autoref{fig:diagram}) generally requires an expert with knowledge of the target language and an understanding of the principles underlying data augmentation. For many low-resource and endangered languages, such experts are rare, and speakers and scholars have many competing obligations. 
In this work, we 
examine whether this expert effort is worthwhile by carefully comparing linguistically-motivated strategies with  strategies that can be implemented without a language expert.

We study a variety of data augmentation methods across two low-resource languages, Arapaho and Uspanteko. We evaluate on translation (to a high-resource language) in both directions and on the task of \textit{interlinear glossing}, where the model generates a sequence of morphological glosses corresponding to the input sentence \cite{ginn-etal-2023-findings}. To design linguistically-motivated augmentation strategies, our first author, a trained linguist, extensively studied linguistic reference materials for both Arapaho and Uspanteko.

We compare non-linguistic augmentation strategies, such as random word insertion or deletion, with our strategies designed to generate grammatically valid examples.
We find that the linguistically-motivated strategies can provide small benefits over the non-linguistic approaches in some cases. However, in cases where the linguistic strategies produce examples which are grammatically valid, but rare or unusual, performance is actually worse for the augmented models. We conclude that while the incorporation of linguistic expert knowledge may be beneficial, it must consider \emph{both} linguistic grammaticality and the target data distribution.

Our specific contributions are the following:
\begin{itemize}
    \item A systematic and comprehensive comparison of various linguistic and non-linguistic data augmentation strategies for low-resource machine translation and interlinear glossing.
    \item Analysis of the effect of combining various augmentation strategies.
    \item Analysis of the interaction between the size of the original training set and the benefits from data augmentation.
\end{itemize}

Our code is available on GitHub\footnote{\url{https://github.com/lecs-lab/is-ling-augmentation-worth-it}} and our results are available on WandB.\footnote{\url{https://wandb.ai/augmorph}}

\section{Related Work}
\subsection{Non-linguistic Augmentation}
Several methods for augmentation have been proposed that do not rely on linguistic knowledge, instead relying on shallow heuristics or statistical methods to produce novel examples, which may or may not be valid utterances in the language. 

\textit{Backtranslation} is a common technique in machine translation, where monolingual data in the target language is translated into the source language \citep{sennrich-etal-2016-improving}, though the resulting examples may not be completely valid. \citet{wang-yang-2015-thats} generate novel sentences by replacing words with other words that have similar static embeddings. Likewise, \citet{fadaee-etal-2017-data} seek to produce valid sentences by substituting words that produce high-probability sentences according to a language model. \citet{andreas-2020-good} perform a similar procedure with sentence fragments, searching for phrases that appear in similar contexts.

\citet{Wei2019} apply word substitutions, deletions, and insertions, performing perturbations that do not necessarily produce valid sentences. \citet{karimi-etal-2021-aeda-easier} use a similar approach but manipulate only punctuation marks. Many additional studies have considered similar heuristics for random perturbation \citep{silfverberg-etal-2017-data, wang-etal-2018-switchout, anastasopoulos-neubig-2019-pushing, guo-etal-2020-sequence, liu-hulden-2022-transformer}.



\subsection{Linguistic Augmentation}
While work in the prior section replaces words or phrases according to statistical patterns, other work proposes the use of linguistic resources to identify valid replacements. This has been done with thesauri \citep{NIPS2015_250cf8b5}, WordNet \citep{Wei2019}, and (for code-switched text) bilingual lexicons  \citep{pratapa-etal-2018-language, winata-etal-2019-code, tarunesh-etal-2021-machine}. Instead of entire words, some research modifies the linguistic features of selected words in each sentence, such as pronominal gender \citep{zhao-etal-2018-gender} or verbal inflection \citep{li-he-2021-data}. Still other research generates entirely synthetic examples by combining morphemes \citep{seo-etal-2023-chef} or sampling from formal grammars such as finite-state machines \citep{lane-bird-2020-bootstrapping} and context-free grammars \citep{lucas-etal-2024-grammar}.


\subsection{Our Contributions}
Our work is novel by providing a careful comparison of similar linguistic and non-linguistic strategies. Additionally, most previous work uses shallow knowledge about the language in the form of dictionaries and thesauri, while we utilize a trained linguist and full reference grammars. 

The closest prior works are \citet{dai-adel-2020-analysis, kashefi-hwa-2020-quantifying}, which compare linguistic and non-linguistic augmentation strategies, but their work studies classification tasks, while we experiment with sequence-to-sequence tasks. Sequence-to-sequence tasks present additional difficulties for effective data augmentation. For classification tasks, it is trivial to ensure the labels for the synthetic data adhere to the set of valid labels; however, for sequence-to-sequence datasets, the labels are unrestricted sequences, and thus it is far more difficult to guarantee their validity.

\section{Datasets and Tasks}
We use the datasets from \citet{ginn-etal-2023-findings}. Each example consists of a sentence in the target language, a translation into Spanish (for Uspanteko) or English (for Arapaho), and a line of \textit{interlinear glosses}. Interlinear glosses provide a tag for each morpheme in the original sentence, which may either be a translation (for stem morphemes) or morphological category. Below are examples for Uspanteko \autoref{ex:usp-gloss} and Arapaho \autoref{ex:arp-gloss}.

\begin{small}
\begin{exe}
  \ex
  \gll  wi' neen tb'ank juntir \\
       EXS INT INC-hacer-SC todo \\
  \glt ``Tienen que hacer todo'' 
  \label{ex:usp-gloss}
\end{exe}
\end{small}

\begin{small}
\begin{exe}
  \ex
  \gll Nihtooneete3eino’ hini’ xouu \\
       PAST-almost-run.into-1S that.those skunk \\
  \glt ``I almost ran into that skunk'' 
  \label{ex:arp-gloss}
\end{exe}
\end{small}
We use a fixed test set, and dynamically create three different evaluation sets by splitting the training set for each random seed. We report the splits in \autoref{tab:splits}. 

\begin{table}[h]
    \centering
    \begin{tabular}{c | c c c}
    \hline
        Language & \# train & \# eval & \# test \\
        \hline
        Uspanteko & 9096 & 479 & 1064 \\
        Arapaho & 41824 & 2202 & 4892 \\
        \hline
    \end{tabular}
    \caption{The number of sentences per dataset split. The test set is fixed across all runs. The eval set is dynamically created across runs by splitting the original training set, to ensure we don't overfit to a particular eval dataset.}
    \label{tab:splits}
\end{table}

The three tasks we study are translation from the target language to a high-resource language (\emph{usp}~$\rightarrow$~\emph{esp}, \emph{arp}~$\rightarrow$~\emph{eng}), translation in the opposite direction (\emph{esp}~$\rightarrow$~\emph{usp}, \emph{eng}~$\rightarrow$~\emph{arp}), and interlinear glossing (\emph{usp}~$\rightarrow$~\emph{igt}, \emph{arp}~$\rightarrow$~\emph{igt}). For the latter, the input is the sentence in the target language (first line) and the desired output is the interlinear gloss line (second line).

\begin{table}[bt]
  \centering
  \begin{tabular}{lcc}
    \hline
    \textbf{Name} & \textbf{Category} & \textbf{\# examples} \\
    \hline
    \multicolumn{3}{l}{\textit{\textbf{Uspanteko}}} \\
    \textsc{Upd-TAM} & Linguistic & 0.3 \\
    \textsc{Ins-Conj} & Linguistic & 20.0\\
    \textsc{Ins-Noise} & Non-linguistic  & 20.0 \\
    \textsc{Del-Any} & Non-linguistic & 0.2 \\
    \textsc{Del-Excl} & Linguistic & 0.2 \\
    \textsc{Dup} & Non-linguistic & 0.3 \\
    \hline
    \multicolumn{3}{l}{\textit{\textbf{Arapaho}}} \\
    \textsc{Ins-Intj} & Linguistic & 20.0 \\
    \textsc{Ins-Noise} & Non-linguistic & 20.0 \\
    \textsc{Perm} & Linguistic & 10.0 \\
  \end{tabular}
  \caption{An overview of the data augmentation methods used in our study. We categorize the strategy as either non-linguistic (random perturbation) or linguistic (linguistically-motivated transformations). In addition, we report the average number of new, synthetic examples created \textit{for each original example}.}
  \label{tab:strategies}
\end{table}

\section{Emulating a Linguistic Expert}
Designing linguistically-motivated augmentation strategies requires in-depth knowledge of linguistics and of the target language. We did not have access to an expert for our target languages, but we emulated this by having our first author extensively study the grammars of the Uspanteko and Arapaho languages. The author has formal training in linguistics at the graduate level, but no prior exposure to Uspanteko or Arapaho. 

In order to gain a strong understanding of the grammars of these languages, the first author spent over a year and nearly 200 hours studying linguistic materials (primarily reference grammars and bilingual dictionaries) and interlinear gloss datasets. The linguistic materials included \citet{Coon2016} and \citet{319093} for Uspanteko and \citet{arp_cowell} for Arapaho. By the end of this period, the first author--while not a fluent speaker of either language--was able to create fully grammatical sentences, following the reference materials.

\section{Augmentation Strategies}
We design both linguistic and non-linguistic augmentation strategies and describe them here (summary in ~\autoref{tab:strategies}).


\subsection{Uspanteko}
Uspanteko is an endangered language spoken in Guatemala with fewer than 6000 speakers \cite{Bennett2022}. Uspanteko is an agglutinating language with complex verbs that may include morphemes for TAM (tense-aspect-mood), person, and other suffixes \cite{Coon2016}. 


We design six augmentation strategies for Uspanteko that include three linguistic and three non-linguistic methods. When running augmentation, we modify the original Uspanteko sentence, as well as the corresponding Spanish sentence (for translation) and interlinear glosses (for IGT generation), adding, deleting, or replacing words as needed. When necessary, we use the Spanish-Uspanteko bilingual dictionary of \citet{319093} to translate words. Examples of each method are shown in \autoref{tab:uspanteko_ex}.

\begin{enumerate}
    \item \textsc{Upd-TAM}: Uspanteko obligatorily marks aspect on the verb, and completive (COM) and incompletive (INC) are high-frequency aspect markers that are easily mapped to their Spanish equivalents. This strategy updates the TAM marker to change completive verbs into incompletive, and vice versa. We skip examples that don't have a verb beginning with COM or INC. To make sure the Spanish translation matches the updated Uspanteko sentence, we use  mlconjug3 \cite{mlconjug3} to update the Spanish verb conjugations.
    
    \item \textsc{Ins-Conj}: Inserts a random conjunction or adverb at the start of the sentence (which is generally valid in Uspanteko), using twenty common conjunctions and adverbs from the \citet{ginn-etal-2023-findings} dataset. 
    \item \textsc{Ins-Noise}: Inserts a random word at the start of the sentence, using twenty random words\footnote{We chose twenty words to match the twenty conjunctions/adverbs in the prior strategy.} from the training data (which are not conjunctions or adverbs). Unlike the prior strategy, this is not guaranteed to produce a linguistically well-formed sentence, allowing us to directly compare whether linguistically-motivated insertion has any benefits over a purely random strategy.
    
    \item \textsc{Del-Any}: Randomly deletes a word from the sentence by index, as well as the corresponding index in the translation and glosses.
    
    \item \textsc{Del-Excl}: Randomly deletes a word from the sentence by index, excluding verbs. If the randomly selected index refers to a verb, the example is skipped and not used for data augmentation. Unlike the prior strategy, this approach seeks to avoid producing entirely ungrammatical sentences.\footnote{
    Both delete strategies are restricted to examples where all four lines of the gloss have the same number of whitespace-separated words, in order to reduce the likelihood of the wrong word being deleted.}
    \item \textsc{Dup}: Duplicates the word at a randomly chosen index.\footnote{Restricted like the previous strategyy.}
    
\end{enumerate}

\begin{table*}[!b]
    \centering
    \begin{tabular}{c|c c c c c}
    \hline
        Task & 100 & 500 & 1000 & 5000 & full \\
        \hline
        \textit{\textbf{Uspanteko}} &  \\
        \emph{usp} $\rightarrow$ \emph{esp} & 14.6 (0.8) & 26.4 (0.1) & 31.7 (0.5) & 44.1 (0.4) & 45.2 (1.9) \\
        \emph{esp} $\rightarrow$ \emph{usp} & 13.7 (0.6) & 23.1 (0.3) & 29.1 (0.7) & 39.6 (0.7) & 40.6 (0.6) \\
        \emph{usp} $\rightarrow$ \emph{igt} & 18.4 (2.0) & 53.9 (1.9) & 65.2 (0.6) & 74.5 (0.8) & 75.4 (0.1) \\
        \hline 
         \textit{\textbf{Arapaho}} &  \\
         \emph{arp} $\rightarrow$ \emph{eng} & 15.3 (0.6)  & 18.7 (0.2)& 22.2 (0.6) & 31.0 (0.4) & 38.9 (0.2) \\
         \emph{eng} $\rightarrow$ \emph{arp} & 21.8 (0.7) & 27.4 (0.2) & 30.7 (0.9) & 40.4 (0.6) & 46.2 (2.3) \\
         \emph{arp} $\rightarrow$ \emph{igt} & 17.7 (1.0) & 38.7 (2.0) & 51.2 (0.6) & 68.0 (0.3) & 76.7 (0.1) \\
    \hline
    \end{tabular}
    \caption{\textbf{Baseline} chrF scores (without any augmented data) across languages, tasks, and training sizes. Reported as the mean over three runs, with the format $mean (std)$.}
    \label{tab:baselines}
\end{table*}

\subsection{Arapaho}
Arapaho is an endangered language spoken in the United States, primarily in Wyoming and Oklahoma, with fewer than 300 fluent speakers \cite{arp_cowell}.  Arapaho is a polysynthetic language with free word order and highly complex verbs \cite{arp_cowell}. Unlike Uspanteko, it is quite difficult to modify verbs in a way that guarantees a valid sentence, so we instead focus on sentence-level augmentation strategies. Examples of each method are shown in \autoref{tab:arapaho_ex}.

\begin{enumerate}
    \item \textsc{Ins-Intj}: Inserts an interjection at the start of the sentence, using twenty common interjections, greetings, and conjunctions from the original textual data. \footnote{There was a limited number of isolatable conjunctions in the data, so we included interjections and greetings in order to have a list of comparable size to the Uspanteko methods. This limitation prevented us from increasing the set size beyond twenty across languages and methods.}
    
    \item \textsc{Ins-Noise}: Similar to the Uspanteko version, inserts a random word at the start of the sentence. The word list is composed of twenty words from the training set. The majority of these are nouns, as they were easiest to isolate and confidently identify.
    
    \item \textsc{Perm}: Produces up to 10 permutations of the original word order.\footnote{We set a limit of 10 to prevent ``drowning'' the model with a potentially huge number of augmented samples.} The permuted sentences are linguistically valid, but may not be preferred by a native speaker due to pragmatic factors \cite{arp_cowell}. 
\end{enumerate}

\section{Experimental Setup}
For each task and language, we run experiments to evaluate the effect of data augmentation on task performance. We train models using the \textsc{ByT5-small} pretrained model \citep{xue-etal-2022-byt5}, a 300 million parameter encoder-decoder transformer model that operates over byte sequences.\footnote{This avoids the issues that come with tokenization and rare languages, and has been shown to be beneficial on these specific datasets \citep{he-etal-2023-sigmorefun}.} The inputs are formatted with a short prompt, such as the example in \autoref{tab:ex_prompt}.

\begin{table}[h]
    \centering
    \begin{tabular}{|l|p{0.7\linewidth}|}
    \hline
        Input & Translate into English: Henee3nee-3oonouh'ut niine'eehek nehe' hotii \\
        Label & This car is very cheap. \\
        \hline
    \end{tabular}
    \caption{An example prompt used for training, in this case to translate from Arapaho into English. We use different prefixes for each of the tasks, though this is likely not strictly necessary.}
    \label{tab:ex_prompt}
\end{table}

As a baseline, we finetune models on the original training set, using the hyperparameters described in \autoref{sec:training_details}. We also finetune models on the augmented sets created by each strategy, using the same hyperparameters.


Theoretical perspectives have claimed that the key to successful data augmentation is creating a diverse set of augmented examples that is not too similar to the original data \citep{feng-etal-2021-survey}. Thus, in addition to individual strategies, we finetune models on each possible combination of two or more augmentation strategies, for a total of $2^6=64$ experimental settings for Uspanteko and $2^3=8$ settings for Arapaho. 

In addition, we wish to disentangle the effects of augmentation at different training sizes. We sample a subset of the training data and use the subset to produce the augmented training set. We experiment with samples of 100, 500, 1000, and 5000 examples, in addition to the full training set. For every setting, we train 3 different models with different random seeds and different subsets at that size. For each task, we train a total of $5 \times 3 \times 64 = 960$ models for Uspanteko and $5 \times 3 \times 8 = 120$ models for Arapaho.

Finetuning is performed with a fixed number of training steps across all settings. We use a learning curriculum where the model is first trained on the synthetic data, followed by the original data, resetting the optimizer in between phases. This approach, used in \citet{lucas-etal-2024-grammar}, essentially treats the augmented data as pretraining data and controls for any effect that might arise with mixing the augmented data into the original dataset. 

\section{Results}
We report the complete set of results for all 1000+ settings on our GitHub,\footnote{\url{https://github.com/lecs-lab/is-ling-augmentation-worth-it}} and highlight the key results here. In addition, for all of the visualizations in this section, we report the results in tabular format in \autoref{sec:numbers}.

In \autoref{tab:baselines}, we report the chrF score\footnote{We chose not to use chrF++ or BLEU as many of our examples are polysynthetic sentences with very few words, so word gram-based methods are less informative.} for the baseline models (no augmentation) across languages and tasks. As expected, the scores are virtually zero for the smallest training size setting, with continual improvements as the amount of training data increases. We also observe that the interlinear glossing task (\emph{usp/arp}~$\rightarrow$~\emph{igt}) is far easier than the translation task, likely because the output sequences are restricted to gloss sequences, which is a smaller output space than translated text. We also observe that translation into the higher-resource language (Spanish or English) is easier than the reverse; one possible explanation is that the pretrained ByT5 model has already been trained to output valid text in those languages, but not in the low-resource languages.


\begin{figure*}[!bt]
    \centering
  \includegraphics[width=0.80\linewidth]{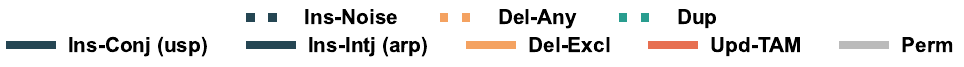} 
  \vspace{10pt}
  \begin{minipage}[b]{0.32\linewidth}
    \includegraphics[width=\linewidth]{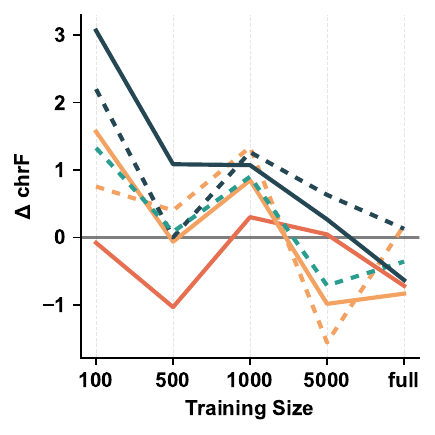}
    \subcaption{\emph{usp} $\rightarrow$ \emph{esp}}
  \end{minipage}
  \hfill
  \begin{minipage}[b]{0.32\linewidth}
    \includegraphics[width=\linewidth]{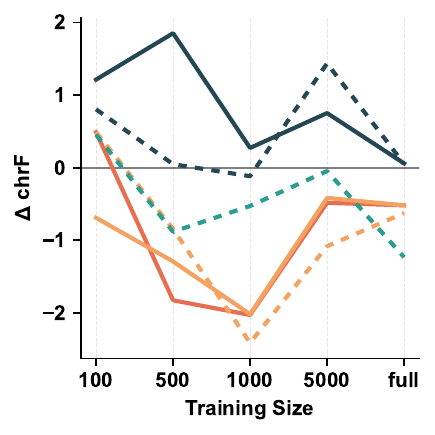}
    \subcaption{\emph{esp} $\rightarrow$ \emph{usp}}
  \end{minipage}
  \hfill
  \begin{minipage}[b]{0.32\linewidth}
    \includegraphics[width=\linewidth]{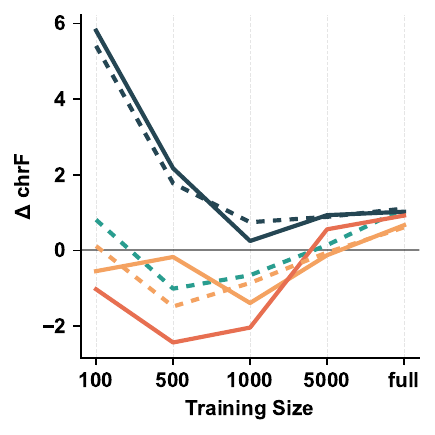}
    \subcaption{\emph{usp} $\rightarrow$ \emph{igt}}
  \end{minipage}

  \begin{minipage}[b]{0.32\linewidth}
    \includegraphics[width=\linewidth]{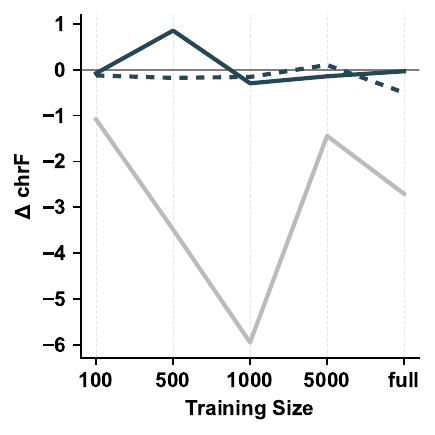}
    \subcaption{\emph{arp} $\rightarrow$ \emph{eng}}
  \end{minipage}
  \hfill
  \begin{minipage}[b]{0.32\linewidth}
    \includegraphics[width=\linewidth]{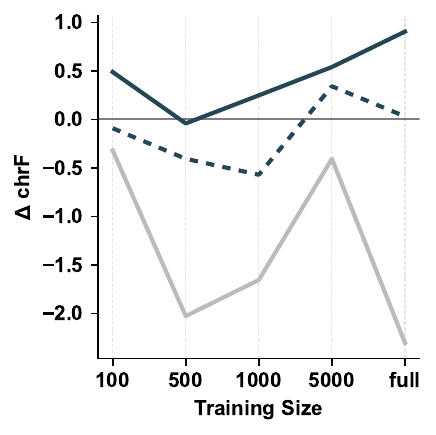}
    \subcaption{\emph{eng} $\rightarrow$ \emph{arp}}
  \end{minipage}
  \hfill
  \begin{minipage}[b]{0.32\linewidth}
    \includegraphics[width=\linewidth]{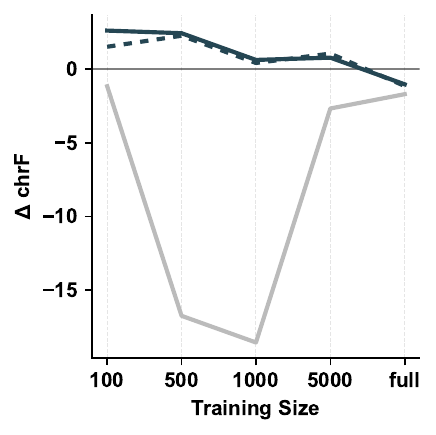}
    \subcaption{\emph{arp} $\rightarrow$ \emph{igt}}
  \end{minipage}

\caption{Difference in (test set) chrF score for various individual augmentation strategies from the baseline (black) for Uspanteko (top) and Arapaho (bottom). Dashed lines indicate non-linguistic strategies, while solid lines are used for linguistic strategies. Averaged over three runs at each point. Tabular form in \autoref{tab:accuracy_ind_scores}.}
    \label{fig:single_results}
\end{figure*}

In \autoref{fig:single_results}, we visualize the performance impact (chrF score) of the individual augmentation strategies as an increase or decrease compared to the baseline performance. We observe that the majority of strategies actually worsen performance somewhat. The only strategies that seem to consistently improve performance are the \textsc{Ins-Noise} and \textsc{Ins-Conj} strategies (in most cases for the latter). 

\begin{figure*}[!bt]
  \centering
  \begin{minipage}[b]{0.32\linewidth}
    \includegraphics[width=\linewidth]{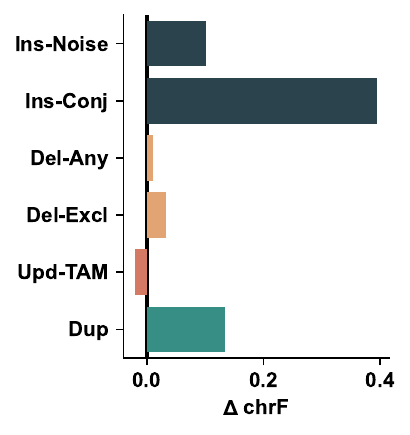}
    \subcaption{\emph{usp} $\rightarrow$ \emph{esp}}
  \end{minipage}
  \hfill
  \begin{minipage}[b]{0.32\linewidth}
    \includegraphics[width=\linewidth]{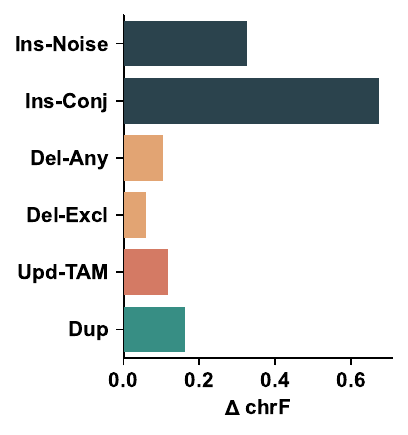}
    \subcaption{\emph{esp} $\rightarrow$ \emph{usp}}
  \end{minipage}
  \hfill
  \begin{minipage}[b]{0.32\linewidth}
    \includegraphics[width=\linewidth]{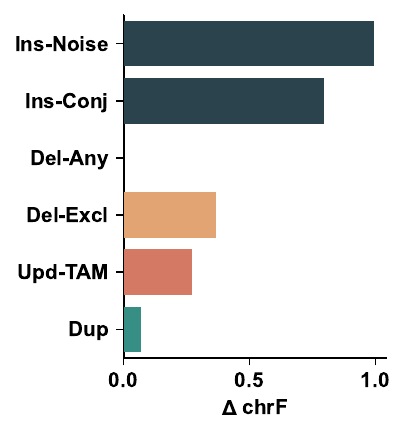}
    \subcaption{\emph{usp} $\rightarrow$ \emph{igt}}
  \end{minipage}

  \begin{minipage}[b]{0.32\linewidth}
    \vspace{5mm}
    \includegraphics[width=\linewidth]{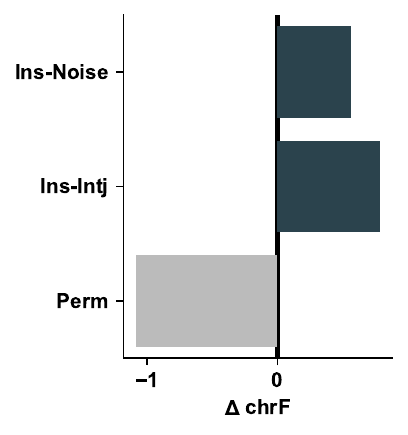}
    \subcaption{\emph{arp} $\rightarrow$ \emph{eng}}
  \end{minipage}
  \hfill
  \begin{minipage}[b]{0.32\linewidth}
    \includegraphics[width=\linewidth]{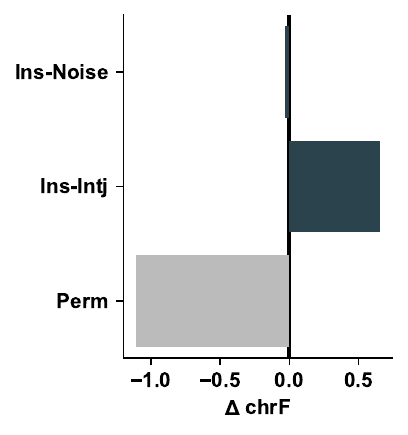}
    \subcaption{\emph{eng} $\rightarrow$ \emph{arp}}
  \end{minipage}
  \hfill
  \begin{minipage}[b]{0.32\linewidth}
    \includegraphics[width=\linewidth]{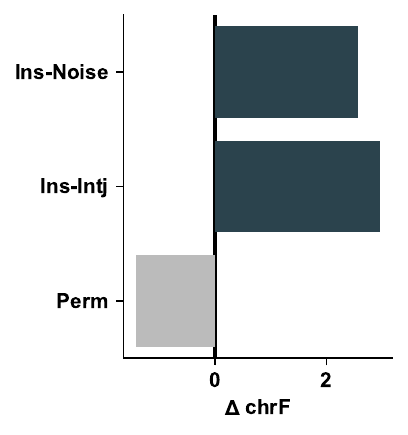}
    \subcaption{\emph{arp} $\rightarrow$ \emph{igt}}
  \end{minipage}

\caption{Average difference in (test set) chrF score between combinations including a given strategy and combinations excluding that strategy. Averaged over all runs and training sizes. Tabular form in \autoref{tab:all_avg_scores}.}
    \label{fig:avg_differences}
\end{figure*}

In \autoref{fig:avg_differences}, we visualize the impact of adding each strategy to a combined augmentation strategy. We compute the mean improvement for each strategy by taking the mean difference in chrF score between combinations with and combinations without the particular strategy. For example, for the strategy \textsc{Ins-Noise}, we would take the mean of the following:

{\small
$$chrF_{\textnormal{Ins-Noise}} - chrF_{\textnormal{Baseline}}$$
$$chrF_{\textnormal{Ins-Noise, Upd-TAM}} - chrF_{\textnormal{Upd-TAM}}$$
$$chrF_{\textnormal{Ins-Noise, Upd-TAM, Del}} - chrF_  {\textnormal{Upd-TAM, Del}}$$
$$...$$}
This allows us to disentangle the effects of adding a particular strategy from the interactions of the other strategies. We report these differences in \autoref{fig:avg_differences}.

\begin{figure*}[!t]
  \centering
    \includegraphics[width=0.95\linewidth]{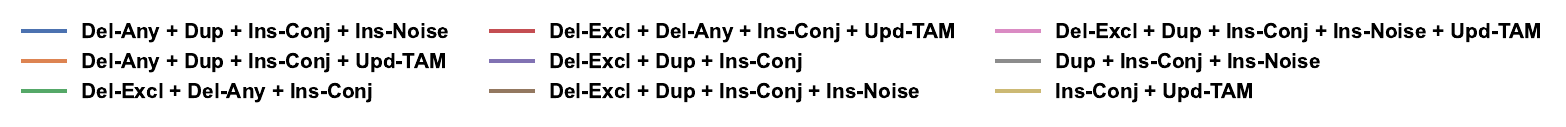}
  \\
  \vspace{10pt}
  \begin{minipage}[b]{0.32\linewidth}
    \includegraphics[width=\linewidth]{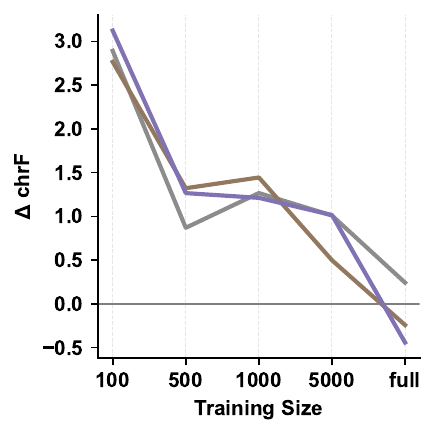}
    \subcaption{\emph{usp} $\rightarrow$ \emph{esp}}
  \end{minipage}
  \hfill
  \begin{minipage}[b]{0.32\linewidth}
    \includegraphics[width=\linewidth]{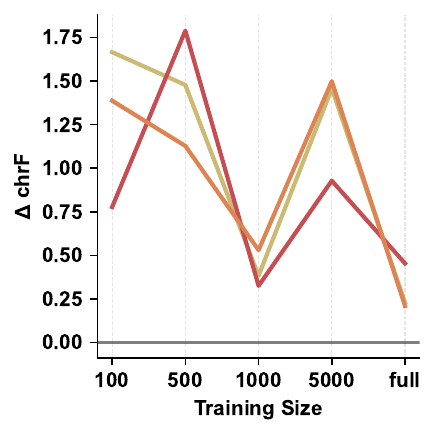}
    \subcaption{\emph{esp} $\rightarrow$ \emph{usp}}
  \end{minipage}
  \hfill
  \begin{minipage}[b]{0.32\linewidth}
    \includegraphics[width=\linewidth]{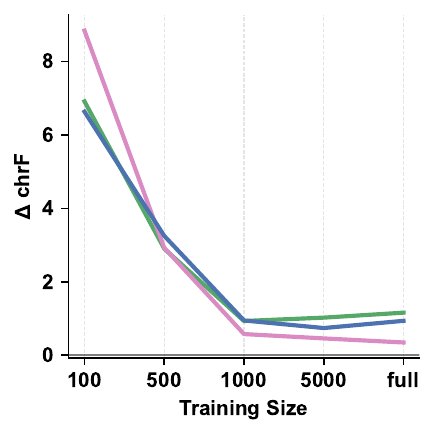}
    \subcaption{\emph{usp} $\rightarrow$ \emph{igt}}
  \end{minipage}

  \vspace{10 pt}
  \includegraphics[width=0.8\linewidth]{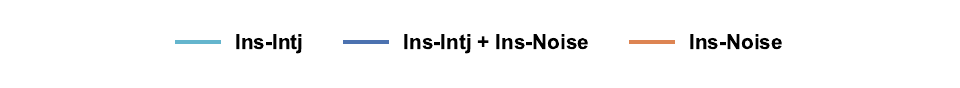}
  \\
  \begin{minipage}[b]{0.32\linewidth}
    \includegraphics[width=\linewidth]{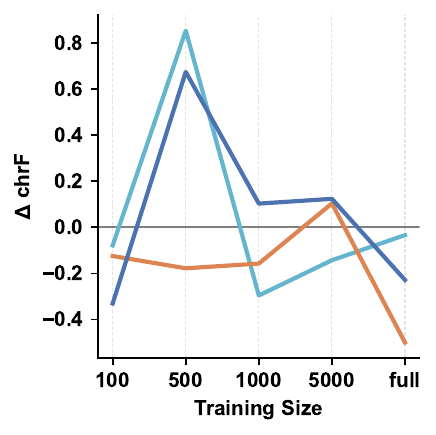}
    \subcaption{\emph{arp} $\rightarrow$ \emph{eng}}
  \end{minipage}
  \hfill
  \begin{minipage}[b]{0.32\linewidth}
    \includegraphics[width=\linewidth]{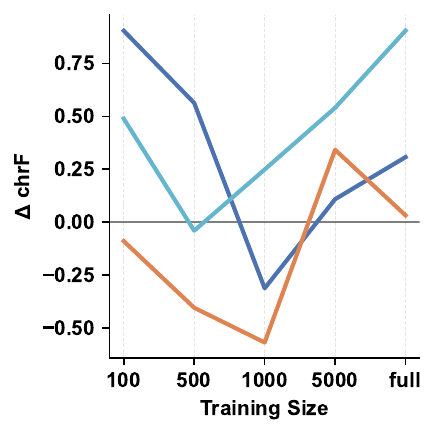}
    \subcaption{\emph{eng} $\rightarrow$ \emph{arp}}
  \end{minipage}
  \hfill
  \begin{minipage}[b]{0.32\linewidth}
    \includegraphics[width=\linewidth]{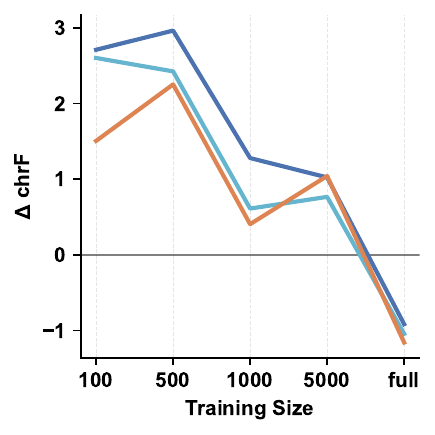}
    \subcaption{\emph{arp} $\rightarrow$ \emph{igt}}
  \end{minipage}
\caption{Performance of the best overall strategies, selected by average performance (chrF score on evaluation set) across training sizes. Performance is reported, as in \autoref{fig:single_results}, as the difference between the target chrF score and the baseline score. Results are averaged over three runs. Tabular form in \autoref{tab:all_accuracy_combined_scores}.}
    \label{fig:combined_results}
\end{figure*}

Finally, in \autoref{fig:combined_results}, we visualize the best overall augmentation strategies for each task, including combined strategies. We compute the mean improvement in chrF scores (on the evaluation set) across the five training set sizes and select the top five augmentation settings. 



\section{Discussion}
All performance improvements are small, with the absolute best strategies achieving an improvement of around +8 (Uspanteko) or +3 (Arapaho) chrF points. This is unsurprising, as these sequence-to-sequence tasks are difficult, and perturbing natural examples may only increase the distributional coverage of our training set by a limited amount. Nonetheless, we do observe improvements on average in \autoref{fig:avg_differences} over all training sizes and combinations. While the improvements from data augmentation alone may not greatly alter the performance of these models, they can certainly be useful in achieving the best possible performance in combination with other techniques.


\paragraph{Effect of linguistically-motivated strategies}
We observe mixed effects from our linguistically-motivated strategies. The \textsc{Upd-TAM} strategy appears to provide small improvements on two of three tasks; however, these improvements are smaller than those of \textsc{Dup}, a completely non-linguistic strategy.

The linguistically-motivated \textsc{Del-Excl} strategy provides a benefit over the corresonding \textsc{Del-Any} for glossing and a small improvement for translation from Uspanteko, but the reverse effect on translation into Uspanteko. It is difficult to interpret this particular result as meaningful evidence one way or another.

On the other hand, we observe a clear improvement of \textsc{Ins-Conj} and \textsc{Ins-Intj} over the corresponding \textsc{Ins-Noise} strategy in both translation tasks (though the IGT task has a smaller or opposite effect). We also observe that \textsc{Ins-Conj} and \textsc{Ins-Intj} are generally the best individual strategies (\autoref{fig:single_results}) and included in most of the top combined strategies (\autoref{fig:combined_results}). In this case, the evidence suggests that the linguistic motivation is beneficial. One possible explanation is that sentences with various conjunctions/interjections as the first word are typical in the actual data distribution, and adding more of such sentences helps the model learn this pattern. On the other hand, sentences with random words inserted in the first position are likely very different from the actual data.

We also observe that the \textsc{Perm} strategy consistently worsens performance, causing over a 1-point drop in chrF for translation (the drop is larger for IGT, which is unsurprising as the gloss sequences depend on the word order of the input sentence). This is an interesting result, as we know that shuffling the word order should always produce a valid sentence in Arapaho (due to the language's free word order). However, even in languages with free word orders, speakers typically exhibit preferences towards certain orderings \cite{dryer1995frequency}. Thus, the augmented examples created by \textsc{Perm} may be very unlike the data distribution, which could account for the clear detriment to performance.

To control for this, we compute an additional evaluation metric. The metric is a variation of chrF that disregards word order, scoring any permutation of the correct words as correct. We compute this by simply removing the character n-grams which cross word boundaries. We report average results for Arapaho \emph{eng}~$\rightarrow$~\emph{arp} setting with chrF and our modified chrF in \autoref{tab:modified_chrf}. We observe that there is a similar trend between the two, indicating that the error is not due to producing sentences with grammatically valid but uncommon word orderings.

\begin{table}[]
    \centering
    \begin{tabular}{l|c c}
    \hline
    & Baseline & \textsc{+Perm} \\
        chrF & 30.0 & 29.0 \\
        Modified chrF &  30.9 & 29.9 \\
        \hline 
    \end{tabular}
    \caption{Average chrF across training sizes for Arapaho \emph{eng}~$\rightarrow$~\emph{arp}, using the standard chrF and a modified chrF that ignores word order. We observe a similar trend regardless of metric.}
    \label{tab:modified_chrf}
\end{table}

These findings point to a key takeaway for linguistically-motivated augmentation: in order to improve performance,\footnote{At least on a held-out test set which is randomly sampled from the same distribution as the training data.} it is not sufficient for the augmented examples to be linguistically valid; they must also be similar (but not too similar) to the target data distribution. On the other hand, it is worth considering whether this is an appropriate evaluation for low-resource translation, since producing grammatical--but unusual--translations may be preferable.

\paragraph{Effect of combined augmentation strategies}
For Uspanteko, we observe that the best overall strategies always include a combination of various augmentation strategies (for Arapaho, there are far fewer possible combinations). One explanation for this is that the use of several strategies produces an augmented dataset with greater diversity, preventing the model from fitting too much to the specific type of augmented example. 

\paragraph{Effect of training set size}
Unsurprisingly we observe that, in most cases, the magnitude of the improvements caused by data augmentation decreases with a larger original training dataset, most dramatically in the \emph{usp}~$\rightarrow$~\emph{esp} and \emph{usp}~$\rightarrow$~\emph{igt} settings. The clear takeaway is that while augmentation can provide benefits in low-resource settings, obtaining additional naturalistic data is more effective. 

\section{Conclusion}
We observe varying performance benefits from different data augmentation strategies on translation and interlinear glossing in two low-resource languages. We consider augmentation strategies which utilize linguistic domain knowledge to produce more linguistically/grammatically valid synthetic examples, and we compare these strategies with approaches that simply utilize random noise and produce potentially ungrammatical examples. We find that the linguistic strategies that match the data distribution most closely (\textsc{Ins-Conj, \textsc{Ins-Intj}}) have clear benefits over the non-linguistic approach. On the other hand, a strategy that produces valid but rare examples (\textsc{Perm}) significantly worsens performance. 

Overall, the answer to our primary research question is cautionary. There do appear to be cases where utilizing linguistic expertise for data augmentation can give an edge over general language-agnostic methods, if the strategies take into account the natural distribution of data. However, the improvements are small, and this may not be the most productive use of expert effort. Instead, this effort could be used to facilitate high-quality data collection and annotation, as collecting additional natural data has clear benefits across NLP tasks.

\section{Limitations}
As several of the methods by their very nature do not target every example in the original dataset (e.g., \textsc{Upd-TAM} is only relevant for sentences containing verbs marked with completive or incompletive aspect), the number of new examples generated varies across strategies. While we control for this effect by using a fixed number of training iterations, it is still possible that having a larger, and thus more diverse, augmented dataset has an effect.


\section{Ethical Considerations}
When working with endangered languages, it is vital to ensure that language data is used in accordance with the wishes of the language community \citep{schwartz-2022-primum}. Furthermore, NLP systems for such languages should be used with caution, as low-quality translation/glossing/etc can be harmful to the language. 

\section*{Acknowledgments}
Author RG was supported in part by funding from CLASIC, CU Boulder's professional MS program in Computational Linguistics. Thanks to the CLASIC cohort and to anonymous reviewers for useful feedback. Parts of this work were supported by the National Science Foundation under Grant No. 2149404, “CAREER: From One Language to Another.” Any opinions, findings, and conclusions or recommendations expressed in this material are those of the authors and do not necessarily reflect the views of the National Science Foundation. This work utilized the Blanca condo computing resource at the University of Colorado Boulder. Blanca is jointly funded by computing users and the University of Colorado Boulder.


\bibliography{anthology,custom}

\appendix
\section{Training Details}
\label{sec:training_details}
We did not perform extensive hyperparameter optimization. For each language, we started with the default parameters and made minor adjustments until we achieved relatively low loss on the training and eval set. We use the Adam optimizer with default parameters and the hyperparameters described in \autoref{tab:hyperparams}. For the augmented models, we first train on the augmented data for the specified number of steps (500 or 2000 for Uspanteko or Arapaho). Then, we train on the original training dataset for 1000 or 4000 steps. For the non-augmented models, we train on the original training dataset in both phases, but still reset the optimizer between phases.

\begin{table}[h]
    \centering
    \begin{tabular}{l|c c}
    \hline
        Parameter & Usp & Arp \\
        \hline 
        Batch size & 32 & 16 \\
        Learning rate & 2E-4 & 2E-4 \\
        Weight decay & 0.5 & 0.5 \\
        Training steps (aug. data) & 500 & 2000 \\
        Training steps (training data) & 1000 & 4000 \\
        \hline 
    \end{tabular}
    \caption{Hyperparameters for all training runs in each language. }
    \label{tab:hyperparams}
\end{table}

The only parameters we specifically tuned were weight decay and the number of training steps, in order to prevent overfitting and ensure convergence. As the Arapaho dataset is roughly four times larger than the Uspanteko dataset, we use four times as many training steps. We train models on several A100 GPUs in the (omitted cluster name for anonymity), and the entire study used around 1000 GPU hours. 

\section{Augmentation Examples}
We provide examples for each strategy in Uspanteko (\autoref{tab:uspanteko_ex}) and Arapaho (\autoref{tab:arapaho_ex}).

\begin{table*}
    \centering
    \caption{Augmentation examples for each method in Uspanteko. Modified parts of the examples are highlighted with bold face.}
    \begin{tabular}{|l|>{\centering\arraybackslash}p{0.7\linewidth}|}
    \hline
        \textsc{Original} &  wi' neen tb'ank juntir \par
                    wi' neen t-b'an-k juntiir\par
                    EXS INT INC-hacer-SC todo\par
                    Tienen que hacer todo\\
        \hline
        \textsc{Upd-TAM} & wi' neen \textbf{x}b'ank juntir\par
                    wi' neen \textbf{x}-b'an-k juntiir\par
                    EXS INT \textbf{COM}-hacer-SC todo\par
                    \textbf{tuvieron} que hacer todo \\
        \hline
        \textsc{Ins-Conj} & \textbf{Pwes} wi' neen tb'ank juntir\par
                    \textbf{Pwes} wi' neen t-b'an-k juntiir\par
                    \textbf{pues} EXS INT INC-hacer-SC todo\par
                    \textbf{Pues }Tienen que hacer todo \\
        \hline
        \textsc{Ins-Noise} & \textbf{Saneb'} wi' neen tb'ank juntir\par
                    \textbf{Saneb'} wi' neen t-b'an-k juntiir\par
                    \textbf{arena@de@rio} EXS INT INC-hacer-SC todo\par
                    \textbf{Harenas del río} Tienen que hacer todo \\
        \hline
        \textsc{Del-Any} &  wi' neen \textbf{[--]} juntir\par
                    wi' neen \textbf{[--]} juntiir\par
                    EXS INT \textbf{[--]} todo\par
                    Tienen que \textbf{[--]} todo \\
        \hline
        \textsc{Del-Excl} &  wi' neen tb'ank \textbf{[--]}\par
                    wi' neen t-b'an-k \textbf{[--]}\par
                    EXS INT INC-hacer-SC \textbf{[--]}\par
                    Tienen que hacer \textbf{[--]}
                     \\
        \hline
        \textsc{Dup} &  wi' neen tb'ank \textbf{tb'ank} juntir\par
                    wi' neen t-b'an-k \textbf{t-b'an-k} juntiir\par
                    EXS INT INC-hacer-SC \textbf{INC-hacer-SC} todo\par
                    Tienen que hacer \textbf{hacer} todo \\
        \hline
    \end{tabular}
    \label{tab:uspanteko_ex}
\end{table*}

\begin{table*}
    \centering
    \caption{Augmentation examples for each method in Arapaho. Modified parts of the examples are highlighted with bold face.}
    \begin{tabular}{|l|>{\centering\arraybackslash}p{0.7\linewidth}|}
    \hline
        \textsc{Original} & Nihtooneete3eino' hini' xouu\par
                    PAST-almost-run.into-1S that(aforementioned).those skunk \par
                    I almost ran into that skunk . \\
        \hline
        \textsc{Ins-Intj} & \textbf{Yeheihoo} Nihtooneete3eino' hini' xouu\par
                    \textbf{gee.whiz }PAST-almost-run.into-1S that(aforementioned).those skunk\par
                    \textbf{Gee whiz }I almost ran into that skunk . \\
        \hline
        \textsc{Ins-Noise} & \textbf{Bih'ih} Nihtooneete3eino' hini' xouu\par
                    \textbf{mule.deer} PAST-almost-run.into-1S that(aforementioned).those skunk\par
                    \textbf{Mule deer} I almost ran into that skunk . \\
        \hline
        \textsc{Perm} & hini' xouu Nihtooneete3eino' \textbf{[order changed]}\par
                    PAST-almost-run.into-1S that(aforementioned).those skunk \textbf{[order changed]}\par
                    I almost ran into that skunk . \textbf{[order changed]}\\
        \hline
    \end{tabular}
    \label{tab:arapaho_ex}
\end{table*}

\section{Table Results}
\label{sec:numbers}
In this section, we provide the corresponding numerical results for all of the visualizations.

\begin{table*}
    \caption{Difference in (test set) chrF score for various individual augmentation strategies from the baseline. Reported as the mean over three runs, with the format mean(std).}
    \label{tab:accuracy_ind_scores}
    \begin{subtable}{\linewidth}
        \caption{\emph{usp} $\rightarrow$ \emph{esp}}
        \label{tab:accuracy_ind_scores_usp_esp}
        \begin{tabular}{p{5cm}| ccccc}
            & 100 & 500 & 1000 & 5000 & full \\
            \hline
            \textsc{Del-Excl} & 1.57 (1.07) & -0.06 (1.27) & 0.85 (0.47) & -0.98 (0.87) & -0.83 (1.99) \\
            \textsc{Ins-Conj} & 3.07 (0.89) & 1.09 (0.43) & 1.07 (1.04) & 0.27 (0.47) & -0.63 (2.04) \\
            \textsc{Ins-Noise} & 2.20 (1.18) & 0.00 (1.28) & 1.27 (0.55) & 0.63 (0.51) & 0.13 (1.81) \\
            \textsc{Del-Any} & 0.76 (0.94) & 0.40 (0.62) & 1.34 (0.56) & -1.56 (0.50) & 0.18 (1.76) \\
            \textsc{Dup} & 1.33 (1.41) & 0.08 (1.24) & 0.90 (0.54) & -0.71 (0.50) & -0.36 (1.79) \\
            \textsc{Upd-TAM}  & -0.08 (0.80) & -1.03 (0.31) & 0.30 (1.19) & 0.05 (0.79) & -0.72 (1.79) \\
            \hline
        \end{tabular}
    \end{subtable}
    \begin{subtable}{\linewidth}
        \vspace{5mm}
        \caption{\emph{esp} $\rightarrow$ \emph{usp}}
        \label{tab:accuracy_ind_scores_esp_usp}
        \begin{tabular}{p{5cm}| ccccc}
            & 100 & 500 & 1000 & 5000 & full \\
            \hline
            \textsc{Del-Excl} & -0.69 (1.01) & -1.29 (0.71) & -2.02 (0.77) & -0.42 (0.59) & -0.52 (1.81) \\
            \textsc{Ins-Conj} & 1.21 (0.88) & 1.85 (0.73) & 0.27 (0.70) & 0.75 (0.82) & 0.06 (0.95) \\
            \textsc{Ins-Noise} & 0.80 (0.81) & 0.05 (1.20) & -0.12 (0.79) & 1.44 (0.63) & 0.05 (0.53) \\
            \textsc{Del-Any} & 0.49 (0.95) & -0.84 (0.98) & -2.41 (1.38) & -1.08 (0.70) & -0.63 (1.19) \\
            \textsc{Dup}  & 0.45 (0.96) & -0.88 (0.39) & -0.53 (1.06) & -0.05 (0.79) & -1.23 (0.59) \\
            \textsc{Upd-TAM}& 0.49 (0.69) & -1.83 (0.73) & -2.03 (1.24) & -0.48 (0.97) & -0.52 (0.90) \\
            \hline
        \end{tabular}
    \end{subtable}
    \begin{subtable}{\linewidth}
        \vspace{5mm}
        \caption{\emph{usp} $\rightarrow$ \emph{igt}}
        \label{tab:accuracy_ind_scores_usp_igt}
        \begin{tabular}{p{5cm}| ccccc}  
            & 100 & 500 & 1000 & 5000 & full \\
            \hline
            \textsc{Del-Excl} & -0.55 (2.77) & -0.17 (1.96) & -1.39 (0.70) & -0.13 (0.78) & 0.67 (0.68) \\
            \textsc{Ins-Conj} & 5.81 (2.74) & 2.17 (1.65) & 0.25 (0.93) & 0.93 (0.77) & 1.02 (0.64) \\
            \textsc{Ins-Noise} & 5.41 (2.93) & 1.78 (1.93) & 0.75 (0.71) & 0.89 (0.80) & 1.11 (0.35) \\
            \textsc{Del-Any}  & 0.12 (1.95) & -1.48 (1.66) & -0.86 (0.75) & -0.05 (0.75) & 0.61 (0.84) \\
            \textsc{Dup} & 0.81 (1.94) & -1.01 (1.75) & -0.66 (0.58) & 0.14 (0.77) & 1.11 (0.20) \\
            \textsc{Upd-TAM}& -1.03 (1.78) & -2.43 (1.66) & -2.04 (0.71) & 0.56 (0.90) & 0.92 (0.51) \\
            \hline
        \end{tabular}
    \end{subtable}
    \begin{subtable}{\linewidth}
        \vspace{5mm}
        \caption{\emph{arp} $\rightarrow$ \emph{eng}}
        \label{tab:accuracy_ind_scores_arp_eng}
        \begin{tabular}{p{5cm}| ccccc}
            & 100 & 500 & 1000 & 5000 & full \\
            \hline
            \textsc{Ins-Intj} & -0.08 (0.97) & 0.85 (0.35) & -0.30 (0.70) & -0.14 (0.87) & -0.03 (0.25) \\
            \textsc{Ins-Noise} & -0.12 (0.62) & -0.18 (0.73) & -0.16 (0.84) & 0.10 (0.87) & -0.50 (0.42) \\
            \textsc{Perm} & -1.08 (0.53) & -3.48 (0.47) & -5.95 (0.58) & -1.44 (0.58) & -2.71 (0.52) \\
            \hline
        \end{tabular}
    \end{subtable}
    \begin{subtable}{\linewidth}
        \vspace{5mm}
        \caption{\emph{eng} $\rightarrow$ \emph{arp}}
        \label{tab:accuracy_ind_scores_eng_arp}
        \begin{tabular}{p{5cm}| ccccc}
            & 100 & 500 & 1000 & 5000 & full \\
            \hline
            \textsc{Ins-Intj} & 0.49 (1.08) & -0.04 (0.79) & 0.25 (0.82) & 0.54 (0.81) & 0.91 (2.20) \\
            \textsc{Ins-Noise} & -0.09 (1.05) & -0.41 (0.87) & -0.57 (0.84) & 0.34 (0.54) & 0.03 (2.43) \\
            \textsc{Perm}  & -0.32 (0.87) & -2.03 (0.93) & -1.66 (0.79) & -0.41 (1.04) & -2.30 (2.87) \\
            \hline
        \end{tabular}
    \end{subtable}
    \begin{subtable}{\linewidth}
        \vspace{5mm}
        \caption{\emph{arp} $\rightarrow$ \emph{igt}}
        \label{tab:accuracy_ind_scores_arp_igt}
        \begin{tabular}{p{5cm}| ccccc}
             & 100 & 500 & 1000 & 5000 & full \\
            \hline
            \textsc{Ins-Intj} & 2.60 (0.91) & 2.43 (1.89) & 0.61 (0.62) & 0.77 (0.82) & -1.04 (0.34) \\
            \textsc{Ins-Noise} & 1.51 (0.93) & 2.25 (2.76) & 0.41 (1.22) & 1.04 (0.48) & -1.16 (1.89) \\
            \textsc{Perm} & -1.17 (1.13) & -16.75 (1.82) & -18.56 (0.76) & -2.68 (0.40) & -1.71 (0.21) \\
            \hline
        \end{tabular}
    \end{subtable}
\end{table*}

\begin{table*}
    \caption{Average difference in (test set) chrF score between combinations including a given strategy and combinations excluding that strategy. Reported as the mean over all runs and training sizes, with the format mean(std).}
    \label{tab:all_avg_scores}
    \begin{subtable}{\linewidth}
        \centering
        \caption{\emph{usp} $\rightarrow$  \emph{esp}}
        \label{tab:avg_scores_usp_esp}
        \begin{tabular}{p{5cm}| c}
            \hline
            \textsc{Ins-Noise} & 0.10 (1.08) \\
            \textsc{Ins-Conj}  & 0.39 (1.17)\\
            \textsc{Del-Any} & 0.01 (0.94)\\
            \textsc{Del-Excl} & 0.03 (0.95) \\
            \textsc{Upd-TAM} & -0.02 (1.00) \\
            \textsc{Dup} & 0.13 (0.98)\\
            \hline
        \end{tabular}
    \end{subtable}
    \begin{subtable}{\linewidth}
        \centering
        \vspace{5mm}
        \caption{\emph{esp} $\rightarrow$  \emph{usp}}
        \label{tab:avg_scores_esp_usp}
        \begin{tabular}{p{5cm}| c}
            \hline
            \textsc{Ins-Noise} & 0.33 (1.21) \\
            \textsc{Ins-Conj} & 0.67 (1.27) \\
            \textsc{Del-Any} & 0.11 (1.06)\\
            \textsc{Del-Excl} & 0.06 (1.10) \\
            \textsc{Upd-TAM} &0.12 (1.08)\\
            \textsc{Dup} & 0.16 (1.08) \\
            \hline
        \end{tabular}
    \end{subtable}
    \begin{subtable}{\linewidth}
        \centering
        \vspace{5mm}
        \caption{\emph{usp} $\rightarrow$ \emph{igt}}
        \label{tab:avg_scores_usp_igt}
        \begin{tabular}{p{5cm}| c}
            \hline
            \textsc{Ins-Noise} & 0.99 (3.59) \\
            \textsc{Ins-Conj} & 0.80 (3.71) \\
            \textsc{Del-Any} & 0.00 (3.30) \\
            \textsc{Del-Excl} & 0.37 (3.29) \\
            \textsc{Upd-TAM}  & 0.27 (3.25) \\
            \textsc{Dup} & 0.07 (3.31) \\
            \hline
        \end{tabular}
    \end{subtable}
    \begin{subtable}{\linewidth}
        \centering
        \vspace{5mm}
        \caption{\emph{arp}$ \rightarrow$ \emph{eng}}
        \label{tab:avg_scores_arp_eng}
        \begin{tabular}{p{5cm}| c}
            \hline
            \textsc{Ins-Noise} & 0.56 (1.45) \\
            \textsc{Ins-Intj} & 0.79 (1.41)\\
            \textsc{Perm} & -1.09 (1.06) \\
            \hline
        \end{tabular}
    \end{subtable}
    \begin{subtable}{\linewidth}
        \centering
        \vspace{5mm}
        \caption{\emph{eng}$ \rightarrow$ \emph{arp}}
        \label{tab:avg_scores_eng_arp}
        \begin{tabular}{p{5cm}| c}
            \hline
            \textsc{Ins-Noise} & -0.03 (1.34) \\
            \textsc{Ins-Intj} & 0.65 (1.56)\\
            \textsc{Perm} & -1.11 (1.78) \\
            \hline
        \end{tabular}
    \end{subtable}
    \begin{subtable}{\linewidth}
        \centering
        \vspace{5mm}
        \caption{\emph{arp}$ \rightarrow$ \emph{igt}}
        \label{tab:avg_scores_arp_igt}
        \begin{tabular}{p{5cm}| c}
            \hline
            \textsc{Ins-Noise} &2.57 (6.00) \\
            \textsc{Ins-Intj} & 2.96 (6.05) \\
            \textsc{Perm} &  -1.43 (0.88)\\
            \hline
        \end{tabular}
    \end{subtable}
\end{table*}

\begin{table*}
    \caption{Performance of the best overall strategies, selected by average performance (chrF score on evaluation set) across training sizes. Performance is reported as the difference between the target chrF score and the baseline score. The results are reported as the mean over three runs, with the format mean(std).}
    \label{tab:all_accuracy_combined_scores}
    \begin{subtable}{\linewidth}
        \caption{\emph{usp} $\rightarrow$  \emph{esp}}
        \label{tab:accuracy_combined_scores_usp_esp}
        \begin{tabular}{>{\centering\arraybackslash}p{5cm}| ccccc}
            & 100 & 500 & 1000 & 5000 & full \\
            \hline
            \textsc{Del-Excl} + \textsc{Dup} + \textsc{Ins-Conj} & 3.12 (0.73) & 1.27 (0.86) & 1.21 (1.29) & 1.02 (0.62) & -0.44 (2.06) \\
            \textsc{Del-Excl} + \textsc{Dup} +\textsc{Ins-Conj} + \textsc{Ins-Noise} & 2.77 (0.78) & 1.32 (0.61) & 1.44 (0.50) & 0.50 (0.61) & -0.24 (1.70) \\
            \textsc{Dup} + \textsc{Ins-Conj} + \textsc{Ins-Noise} & 2.89 (1.45) & 0.87 (1.05) & 1.27 (0.79) & 1.01 (0.42) & 0.24 (1.88) \\
            \hline
        \end{tabular}
    \end{subtable}
    \begin{subtable}{\linewidth}
        \vspace{5mm}
        \caption{\emph{esp} $\rightarrow$ \emph{usp}}
        \label{tab:accuracy_combined_scores_esp_usp}
        \begin{tabular}{>{\centering\arraybackslash}p{5cm}| ccccc}
             & 100 & 500 & 1000 & 5000 & full \\
            \hline
            \textsc{Del-Any} + \textsc{Dup} +\textsc{Ins-Conj} +  \textsc{Upd-TAM} & 1.39 (1.18) & 1.13 (1.47) & 0.53 (0.74) & 1.50 (0.77) & 0.21 (1.00) \\
            \textsc{Del-Excl} + \textsc{Del-Any} + \textsc{Ins-Conj} + \textsc{Upd-TAM} & 0.78 (0.86) & 1.79 (0.61) & 0.32 (1.07) & 0.93 (0.63) & 0.45 (0.83) \\
            \textsc{Ins-Conj} + \textsc{Upd-TAM} & 1.67 (0.78) & 1.48 (0.75) & 0.38 (0.88) & 1.46 (0.68) & 0.23 (0.60) \\
            \hline
        \end{tabular}
    \end{subtable}
    \begin{subtable}{\linewidth}
        \vspace{5mm}
        \caption{\emph{usp}$ \rightarrow$ \emph{igt}}
        \label{tab:accuracy_combined_scores_usp_igt}
        \begin{tabular}{>{\centering\arraybackslash}p{5cm}| ccccc}
            & 100 & 500 & 1000 & 5000 & full \\
            \hline
            \textsc{Del-Any} + \textsc{Dup} +\textsc{Ins-Conj} + \textsc{Ins-Noise} & 6.63 (2.47) & 3.25 (1.74) & 0.95 (0.75) & 0.74 (0.72) & 0.94 (0.29) \\
            \textsc{Del-Excl} + \textsc{Del-Any} +\textsc{Ins-Conj} & 6.91 (2.44) & 2.90 (1.85) & 0.93 (0.58) & 1.03 (0.76) & 1.16 (0.13) \\
            \textsc{Del-Excl} + \textsc{Dup} + \textsc{Ins-Conj} + \textsc{Ins-Noise} + \textsc{Upd-TAM} & 8.84 (3.67) & 2.93 (1.72) & 0.58 (0.59) & 0.46 (0.96) & 0.35 (0.51) \\
            \hline
        \end{tabular}
    \end{subtable}
    \begin{subtable}{\linewidth}
        \vspace{5mm}
        \caption{\emph{arp}$ \rightarrow$ \emph{eng}}
        \label{tab:accuracy_combined_scores_arp_eng}
        \begin{tabular}{p{5cm}| ccccc}
            & 100 & 500 & 1000 & 5000 & full \\
            \hline
            \textsc{Ins-Intj} & -0.08 (0.97) & 0.85 (0.35) & -0.30 (0.70) & -0.14 (0.87) & -0.03 (0.25) \\
            \textsc{Ins-Intj} +  \textsc{Ins-Noise} & -0.33 (0.64) & 0.67 (0.25) & 0.10 (0.50) & 0.12 (0.95) & -0.23 (0.21) \\
            \textsc{Ins-Noise} & -0.12 (0.62) & -0.18 (0.73) & -0.16 (0.84) & 0.10 (0.87) & -0.50 (0.42) \\
            \hline
        \end{tabular}
    \end{subtable}
    \begin{subtable}{\linewidth}
        \vspace{5mm}
        \caption{\emph{eng}$ \rightarrow$ \emph{arp}}
        \label{tab:accuracy_combined_scores_eng_arp}
        \begin{tabular}{p{5cm}| ccccc}
            & 100 & 500 & 1000 & 5000 & full \\
            \hline
            \textsc{Ins-Intj} & 0.49 (1.08) & -0.04 (0.79) & 0.25 (0.82) & 0.54 (0.81) & 0.91 (2.20) \\
            \textsc{Ins-Intj} + \textsc{Ins-Noise} & 0.90 (0.96) & 0.56 (0.66) & -0.31 (1.26) & 0.11 (0.61) & 0.31 (2.43) \\
            \textsc{Ins-Noise} & -0.09 (1.05) & -0.41 (0.87) & -0.57 (0.84) & 0.34 (0.54) & 0.03 (2.43) \\
            \hline
        \end{tabular}
    \end{subtable}
    \begin{subtable}{\linewidth}
        \vspace{5mm}
        \caption{\emph{arp}$ \rightarrow$ \emph{igt}}
        \label{tab:accuracy_combined_scores_arp_igt}
        \begin{tabular}{p{5cm}| ccccc}
            & 100 & 500 & 1000 & 5000 & full \\
            \hline
            \textsc{Ins-Intj} & 2.60 (0.91) & 2.43 (1.89) & 0.61 (0.62) & 0.77 (0.82) & -1.04 (0.34) \\
            \textsc{Ins-Intj} + \textsc{Ins-Noise} & 2.71 (1.20) & 2.97 (2.23) & 1.28 (0.67) & 1.02 (0.56) & -0.91 (0.67) \\
            \textsc{Ins-Noise} & 1.51 (0.93) & 2.25 (2.76) & 0.41 (1.22) & 1.04 (0.48) & -1.16 (1.89) \\
            \hline
        \end{tabular}
    \end{subtable}
\end{table*}

\end{document}